\documentclass{article}
\usepackage{ijcai22}

\usepackage{times}
\usepackage{soul}
\usepackage{url}
\usepackage[hidelinks]{hyperref}
\usepackage[utf8]{inputenc}
\usepackage[small]{caption}
\usepackage{graphicx}
\usepackage{amsmath}
\usepackage{amsthm}
\usepackage{booktabs}
\usepackage{multirow}
\usepackage{caption}
\usepackage{subcaption}
\usepackage{tabularx}
\usepackage{graphicx}
\usepackage{adjustbox}
\usepackage{amsfonts}
\usepackage{fixmath}

\usepackage{algorithm}
\usepackage[noend]{algpseudocode}

\title{Tsetlin Machine Embedding: Representing Words Using Logical Expressions}

\author{
Bimal Bhattarai 
\and
Ole-Christoffer Granmo \and
Lei Jiao \and
Rohan Yadav \And Jivitesh Sharma
\affiliations
Centre for AI Research (CAIR), University of Agder, Norway
\emails
 \{bimal.bhattarai, ole.granmo, lei.jiao, rohan.yadav, jivitesh.sharma\}@uia.no
}

\begin{document}

\maketitle

\begin{abstract}
Embedding words in vector space is a fundamental first step in state-of-the-art natural language processing (NLP). Typical NLP solutions employ pre-defined vector representations to improve generalization by co-locating similar words in vector space. For instance, Word2Vec is a self-supervised predictive model that captures the context of words using a neural network. Similarly, GLoVe is a popular unsupervised model incorporating corpus-wide word co-occurrence statistics. Such word embedding has significantly boosted important NLP tasks, including sentiment analysis, document classification, and machine translation. However, the embeddings are dense floating-point vectors, making them expensive to compute and difficult to interpret. In this paper, we instead propose to represent the semantics of words with a few defining words that are related using propositional logic. To produce such logical embeddings, we introduce a Tsetlin Machine-based autoencoder that learns logical clauses self-supervised. The clauses consist of contextual words like ``black,'' ``cup,'' and ``hot'' to define other words like ``coffee,'' thus being human-understandable. We evaluate our embedding approach on several intrinsic and extrinsic benchmarks, outperforming GLoVe on six classification tasks. Furthermore, we investigate the interpretability of our embedding using the logical representations acquired during training. We also visualize word clusters in vector space, demonstrating how our logical embedding co-locate similar words.\footnote{The Tsetlin Machine Autoencoder and logical word embedding implementation is available here: \href{https://github.com/cair/tmu}{https://github.com/cair/tmu}.}
\end{abstract}

\section{Introduction}
The success of natural language processing (NLP) relies on advances in word, sentence, and document representation. By capturing word semantics and similarities, such representations boost the performance of downstream tasks~\cite{borgeaud2022improving}, including clustering, topic modelling~\cite{angelov2020top2vec}, searching, and text mining~\cite{huang2020embedding}.

While straightforward, the traditional bag-of-words  encoding does not consider the words' position, semantics, and context within a document. Distributed word representation~\cite{bengio2000neural,bojanowski2017enriching} addresses this lack by encoding words as low-dimensional vectors, referred to as \emph{embeddings}. The purpose is to co-locate similar or contextually relevant words in vector space. There are many algorithms for learning word embeddings. Contemporary self-supervised techniques like Word2Vec~\cite{Mikolov2013EfficientEO}, FastText~\cite{bojanowski2017enriching}, and GloVe~\cite{pennington2014glove} have demonstrated how to build embeddings from word co-occurrence, utilizing massive training data. Introducing context-dependent embeddings, the more sophisticated language models BERT~\cite{Devlin2019BERTPO} and ELMO~\cite{peters1802deep} now perform remarkably well in downstream tasks~\cite{reimers2019sentence}. However, they require significant computation power~\cite{Schwartz2020GreenA}. \par

The above approaches represent words as dense floating point vectors. Word2Vec, for instance, typically builds a 300-dimensional vector per word. The size and density of these vectors make them expensive to compute and difficult to interpret. Consider, for example, the word ``queen.'' Representing it with 300 floats seems inefficient compared to the Oxford Language definition for the same word: ``the female ruler of an independent state, especially one who inherits the position by right of birth.'' From this perspective, it appears advantageous to create embeddings directly from words rather than from arbitrary floating-point values.
Such interpretable embeddings would capture the multiple meanings of a word using a few defining words, simplifying both computation and interpretation.

In this paper, we propose a Tsetlin Machine~(TM) \cite{granmo2018tsetlin} based autoencoder for creating interpretable embeddings. The autoencoder builds propositional logic expressions with context words that identify each target word. The term ``coffee'' can, for instance, be represented by ``one,'' ``hot,'' ``cup,'' ``table,'' and ``black.''  In this manner, the TM builds contextual representations from a vast text corpus, which model the semantics of each word. In contrast to neural network-based embedding, the logical TM embedding is sparse. The embedding space consists of, e.g., $500$ truth values, where each truth value is a logical expression over words. For contextual representation, each target word links to less than ten percent of these expressions. Despite the sparsity and crispness of this representation, it is competitive with neural network-based embedding.

The contributions of our work are summarized below:
\begin{itemize}
    \item We propose the first TM-based Autoencoder to learn efficient encodings in a self-supervised manner.
    \item We introduce TM-based word embedding that builds human-comprehensible contextual representations from unlabeled data. 
    \item We compare our embedding with state-of-the-art approaches on several intrinsic and extrinsic benchmarks, outperforming GloVe on six downstream classification tasks.
\end{itemize}

\section{Related Work}

The majority of self-supervised embedding approaches produce dense word representations based on the distributional hypothesis~\cite{harris1954distributional}, which states that words that occur in the same context are likely to have similar meaning. Word2Vec~\cite{Mikolov2013EfficientEO} is one of the best-known models. It builds embeddings from word co-occurrence using a neural network, leveraging the hidden layer output weights.  GloVe~\cite{pennington2014glove}, on the other hand, embeds by factorizing a word co-occurrence matrix. Similarly, canonical correlation analysis~(CCA) is used in~\cite{dhillon2015eigenwords} for embedding words to maximise context correlation. In~\cite{levy2015improving}, it is demonstrated how precise factorization-based SVD can compete with neural embedding. However, all of these methods are challenging to train because they involve tweaking algorithms and hyperparameters toward particular applications~\cite {lample2016neural}, limiting their wider applicability.

\par
Building upon word embedding, several studies focus on sentence embedding~\cite{arora2017simple,logeswaran2018efficient}. Recent advances in sentence embedding include supervised data inference~\cite{reimers2019sentence}, multitask learning~\cite{cer2018universal}, contrastive learning~\cite{zhang2020unsupervised}, and pretrained large language models~\cite{li2020sentence}. However, the majority of sentence embedding techniques overlook intrinsic evaluations such as similarity tasks, and instead largely focus on extrinsic evaluations involving downstream performance. The most recent building block for embedding  originates from the transformer approach~\cite{vaswani2017attention}. Transformers provide context awareness by utilizing stacks of self-attention layers. BERT~\cite{kenton2019bert}, for instance, employs the transformer architecture to carry out extensive self-supervised training, making it capable of producing text embedding. Other embedding models use a contrastive loss function to perform supervised fine-tuning on positive and negative text pairs~\cite{wang2021tsdae}. Despite the large variety of text embedding models, they all share three main drawbacks: i) they are computationally demanding to train; ii) they are intrinsically complex because they are trained on a large amount of data to tune a huge amount of parameters; and iii) the embeddings produced from these models are not easily interpreted by humans. \par
To improve interpretability, Faruqui et al. introduced ``Sparse Overcomplete Word Vectors"~(SPOWV) which create a sparse non-negative projection of word embedding using dictionary learning~\cite{faruqui2015sparse}. Similarly, SParse Interpretable Neural Embeddings~(SPINE) employs a k-sparse denoising autoencoder to generate sparse embeddings~\cite{subramanian2018spine}. However, these methods are unable to distinguish between multiple context-dependent word meanings. To address this problem, another avenue of research focuses on composing linear combinations of dense vectors from Word2Vec and GloVe~\cite{arora2018linear}. However, the assumption of linearity does not hold for real-world data, yielding linear coefficients that are difficult to comprehend~\cite{mu2017geometry}.\par

The logical embedding approach we present here is most closely related to Naive Bayes word sense induction and topic modeling~\cite{charniak2013naive,lau2014learning}. This approach learns word meanings from local contexts by considering each instance of the word in a document as a pseudo-document. However, the approach is not scalable because it requires training a single topic per target word. Our approach, on the other hand, is scalable and builds non-linear (non-naive) logical embeddings that capture word compositions. To build the logical embeddings, we propose a novel human-interpretable algorithm based on the TM that provides logical rules describing contexts. The TM has recently performed competitively with other deep learning techniques in many NLP tasks, including novelty detection~\cite{bhattarai2022tsetlin}, sentiment analysis~\cite{6}, and fake news detection~\cite{8}. Furthermore, the local and global interpretability of TMs have been explored through direct manipulation of the logical rules~\cite{blakely2021closed}.

\section{Tsetlin Machine Autoencoder}\label{sec:tm_autoencoder}

\begin{figure}
\centerline{\includegraphics[width=0.35\textwidth]{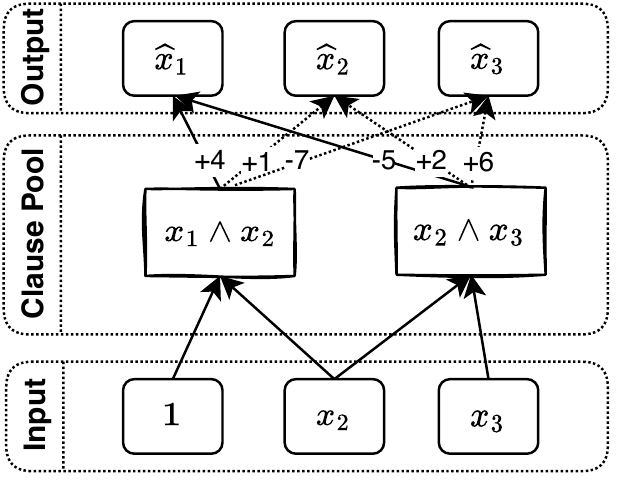}}
\caption{Tsetlin Machine Autoencoder. In this illustration, $x_1$ is masked by
replacing it with value 1 for inferring $\hat{x}_1$.}
\label{figure:tm_autoencoder}
\end{figure}

We here detail the TM Autoencoder based on the Coalesced TM~\cite{glimsdal2021coalesced}, extended with input masking and freezing of masked variables. For ease of explanation, we use three inputs. Adding more inputs follows trivially.

\subsection{Architecture}
\paragraph{Input and Output.} As seen in Figure~\ref{figure:tm_autoencoder}, the TM Autoencoder digests and outputs propositional values: $(x_1, x_2, x_3) \in \{0,1\}^3 \rightarrow (\widehat{x}_1, \widehat{x}_2, \widehat{x}_3) \in \{0,1\}^3$. For our purposes, the propositional variables $x_1$, $x_2$, and $x_3$ each represents a word, for example, ``Brilliant,'' ``Actor,'' and ``Awful.'' The value $1$ means that the word occurs in the input text, while the value $0$ means that it does not. I.e., we represent natural language text as a \emph{set of words}. Notice also that the input variables have corresponding output variables $\widehat{x}_1$, $\widehat{x}_2$, and $\widehat{x}_3$. In short, $\widehat{x}_1$ is to be predicted from $x_2$ and $x_3$, $\widehat{x}_2$ from $x_1$ and $x_3$, and so on. Continuing our example, $\widehat{x}_1$ predicts the presence of ``Brilliant'' based on knowing the occurrence of ``Actor'' and ``Awful.''

\paragraph{Clause Pool.} A pool of $n$ conjunctive clauses, denoted $C_j, j \in \{1, 2, \ldots, n\},$ encodes the input in order to predict the output. A conjunctive clause $C_j$ is simply an \emph{And}-expression over a given subset $L_j \subseteq \{x_1, x_2, x_3\}$ of the input (our autoencoder does not use the input negations $\lnot x_1$, $\lnot x_2$, and $\lnot x_3$):
\begin{equation}
C_j(x_1, x_2, x_3) = \bigwedge_{x_k \in  L_j} x_k.
\end{equation}
For example, the input subset $L_1 = \{x_1, x_2\}$ gives the clause $C_1(x_1, x_2, x_3) = x_1 \land x_2$ in the figure. This clause matches the input if $x_1$ and $x_2$ both are~$1$. In our example, the clause accordingly encodes the concept ``Brilliant Actor''.
\paragraph{Weights.} An integer weight matrix $\mathbold{W}$ connects each of the $n$ clauses to the three outputs $\widehat{x}_1$, $\widehat{x}_2$, and $\widehat{x}_3$:
\begin{equation} \mathbold{W} = \begin{bmatrix} w_{11}&\cdots&w_{1n}\\
w_{21}&\cdots&w_{2n}\\
w_{31}&\cdots&w_{3n}\end{bmatrix} \in \mathbb{Z}^{3 \times n}.
\end{equation}
The row index is an output while the column index is a clause. The weight $w_{12}$, for instance, connects output $\widehat{x}_1$ to clause $C_2$. In Figure \ref{figure:tm_autoencoder}, six weights connect two clauses and three outputs: 
\begin{equation} 
\begin{bmatrix}+4&-5\\+1&+2\\
-7&+6\end{bmatrix}.
\end{equation}
 Consider, for example, the weights $(+4, -5)$ of output $\widehat{x}_1$ in the figure. The weight $+4$ states that clause $C_1(x_1, x_2, x_3) = x_1 \land x_2$ favours $\widehat{x}_1$ being $1$, while clause $C_2(x_1, x_2, x_3) = x_2 \land x_3$ opposes it. For example, the concept ``Awful Actor'' opposes output ``Brilliant.''

\subsection{Inference}
Let us consider the prediction of $\widehat{x}_1$ first. The autoencoder predicts
$\widehat{x}_1$ from the clauses and weights:
\begin{equation}
\widehat{x}_1 = 0 \le \sum_{j=1}^n w_{j1} C_j(1, x_2, x_3).\label{eqn:predict_x1}
\end{equation}
That is, each clause $C_j$ is multiplied by its weight $w_{j1}$ for output  $\widehat{x}_1$. The outcomes are then  summed up to decide the output. If the sum is larger than or equal to zero, the output is $\widehat{x}_1=1$. Otherwise, it is $\widehat{x}_1=0$. Clauses with positive weight thus promote output $\widehat{x}_1=1$ while clauses with negative weight encourage $\widehat{x}_1=0$. Notice that $x_1$ is masked by replacing it with value~$1$. Accordingly, the autoencoder infers output $\widehat{x}_1$ from the remaining inputs $x_2$ and $x_3$.

Correspondingly, $\widehat{x}_2$ and $\widehat{x}_3$  are calculated by respectively masking $x_2$ and $x_3$:
\begin{eqnarray}
\widehat{x}_2&=&0 \le \sum_{j=1}^n w_{j2} C_j(x_1, 1, x_3),\\
\widehat{x}_3&=&0 \le \sum_{j=1}^n w_{j3} C_j(x_1, x_2, 1).
\end{eqnarray}

\paragraph{Example.}  Assume that the input is always either $(1, 1, 0)$ or $(0, 1, 1)$. The input $(1, 1, 0)$ could for instance represent ``Brilliant Actor'' and $(0, 1, 1)$ ``Awful Actor.'' Then notice how Eq.~(\ref{eqn:predict_x1}) correctly determines the masked input $x_1$ with output $\widehat{x}_1$ in Figure~\ref{figure:tm_autoencoder}, both for input $(1, 1, 0)$ and $(0, 1, 1)$.

\begin{figure}
\centerline{\includegraphics[width=0.25\textwidth]{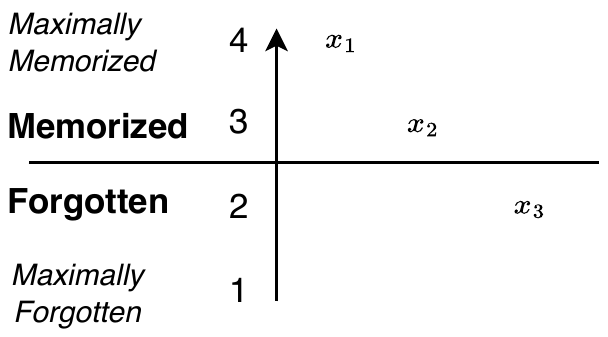}}
\caption{Tsetlin Machine memory for single clause.}
\label{figure:tm_memory}
\end{figure}

\subsection{Learning}
We next consider how to learn the variable subsets $L_j$ for the clauses $C_j, j \in \{1, 2, \ldots, n\},$ as well as how to determine the weights $w_{ji}$ of the weight matrix $W$. 
\paragraph{Clause Memory.} Each clause $C_j$ has a graded memory that contains the input variables, shown in Figure \ref{figure:tm_memory}. The graded memory enables \emph{incremental} learning of the variable subsets from data. Observe how each variable is in one of four memory positions (the number of memory positions is a user-configurable parameter). Positions $1-2$ means Forgotten.  Positions $3-4$ means Memorized. Memorized variables take part in the clause, while Forgotten ones do not. The memory in Figure \ref{figure:tm_memory} thus gives the clause $C_j(x_1, x_2, x_3) = x_1 \land x_2$.

\paragraph{Learning Step.} The TM Autoencoder learns incrementally using three kinds of memory and weight updates: Type~Ia, Type~Ib, and Type~II. Each training example has the form $[k, (x_1, x_2, x_3), x_k], 1 \le k \le 3$. The first element is an index that identifies which input to mask and which output to predict. The second element is an input vector $(x_1, x_2, x_3)$ and the third element is the target value for output $\widehat{x}_k$, which is $x_k$.  We describe the update procedure step-by-step below for index $1$ examples (output $\widehat{x}_1$ prediction). The update procedure for $\widehat{x}_2$ and $\widehat{x}_3$ follows trivially. 

\paragraph{Clause Update Probability.} First, we calculate the weighted clause sum for $\widehat{x}_1$ from Eqn. (\ref{eqn:predict_x1}): $v_1 = \sum_{j=1}^n w_{j1} C_j(1, x_2, x_3)$. The sum is then compared with a margin $T$ (hyper-parameter) to calculate a summation error $\epsilon$. The error depends on the $x_1$-value:
\begin{equation}
\epsilon = \begin{cases}
T - \mathrm{clip}(v_1, -T, T),& x_1=1,\\
T + \mathrm{clip}(v_1, -T, T),& x_1=0.
\end{cases}
\end{equation}
That is, for $x_1$-value $1$ the weighted clause sum should become $T$, while for $x_1$-value $0$ the sum should become $-T$. The goal of the learning is thus to reach the margin for all inputs $(x_1, x_2, x_3)$, ensuring correct output from Eqn. (\ref{eqn:predict_x1}).  To reach this goal, each clause $C_j$ is updated randomly with probability $\frac{\epsilon}{2T}$ in each round. In other words, the update probability drops with the error towards zero.

\paragraph{Update Types.} The kind of update depends on the values of $x_1$, $C_j(1, x_2, x_3)$, and $w_{j1}$. We first consider clauses with positive weight, $w_{j1} \ge 0$. According to Eqn. \ref{eqn:predict_x1}, they are to recognize patterns for $x_1=1$. Note that in all of the below updates, the masked variable $x_1$ is frozen, leaving it unaffected by the update.

\begin{figure}
\centerline{\includegraphics[width=0.25\textwidth]{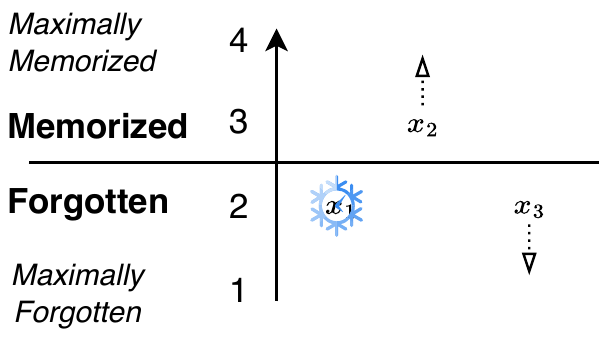}}
\caption{Type Ia (Recognize) Feedback for input $(1, 1, 0)$. The masked variable $x_1$ is frozen.}
\label{figure:tm_memory_recognize}
\end{figure}

\begin{itemize}
\item \textbf{Type Ia (Recognize) Feedback} occurs when $x_1=1$ and $C_j(1, x_2, x_3)=1$. Then one can say that $C_j(1, x_2, x_3)=1$ is a \emph{true positive} because it correctly predicts the masked $x_1$-value. The Type Ia feedback reinforces this successful match by updating the memory of $C_j$ to further mimic the input (see Figure \ref{figure:tm_memory_recognize}). That is, $1$-valued variables move one step upwards in memory, with probability $1.0$.\footnote{Originally, the increment probability is  $\frac{s-1}{s}$, which can be boosted to $1.0$ to enhance learning of true positive patterns~\cite{granmo2018tsetlin}.} Conversely, $0$-valued inputs move one step downwards, however, randomly with probability $\frac{1}{s}$. Here, $s$ is a hyperparameter called \emph{specificity}, meaning that  a larger $s$ makes the clauses more specific. The clause overall is also reinforced by incrementing its weight~$w_{j1}$ by $1$.
\begin{figure}
\centerline{\includegraphics[width=0.25\textwidth]{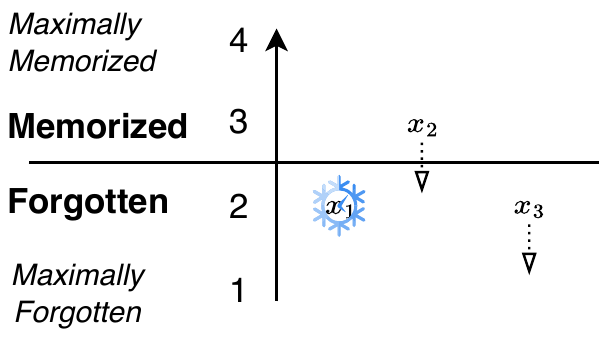}}
\caption{Type Ib (Erase) Feedback for input $(0, 0, 1)$. The masked variable $x_1$ is frozen.}
\label{figure:tm_memory_erase}
\end{figure}
\item \textbf{Type Ib (Erase) Feedback} occurs when $x_1=1$ and $C_j(1, x_2, x_3)=0$. Then we call $C_j(1, x_2, x_3)=0$ a \emph{false negative} because it fails to promote $x_1=1$. In that case, all inputs randomly move one step downwards in memory (see Figure \ref{figure:tm_memory_erase}). Again, each downward move happens with probability~$\frac{1}{s}$. Here, the purpose is to eliminate the false negative outcome by erasing variables from the clause.
\begin{figure}
\centerline{\includegraphics[width=0.25\textwidth]{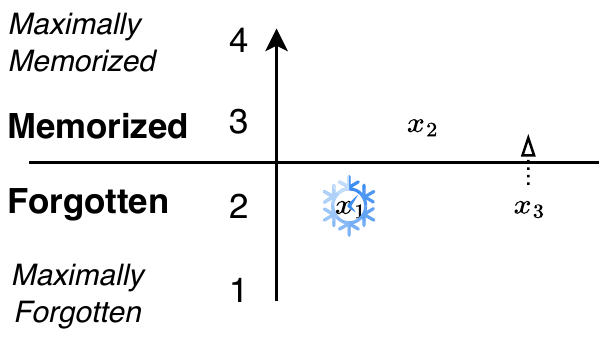}}
\caption{Type II (Reject) Feedback for input $(0, 1, 0)$. The masked variable $x_1$ is frozen.}
\label{figure:tm_memory_reject}
\end{figure}
\item \textbf{Type II (Reject) Feedback} occurs when $x_1=0$ and $C_j(1, x_2, x_3)=1$. Then, one can say that $C_j(1, x_2)=1$ is a \emph{false positive} because it promotes $x_1=1$ when in fact we have $x_1=0$. Then all Forgotten $0$-valued inputs move one step upwards in memory. The purpose is to eventually eliminate the current false positive outcome by injecting $0$-valued variables into the clause. The clause is further diminished by decrementing its weight~$w_{j1}$ by $1$. Note that the latter decrement can switch the weight from positive to negative. In effect, the clause then changes role, now training to recognize $x_1=0$ instead.
\end{itemize}
Clauses $C_j$ with negative weights, $w_{j1} < 0$, are updated the same way. However, they are to recognize patterns for $x_1=0$. To achieve this, $x_1=0$ is treated as $x_1=1$ and $x_1=1$ is treated as $x_1=0$ when updating the memories. Furthermore, the weight updates are reversed. Increments becomes decrements, and vice versa.

\begin{algorithm}
\small
\caption{TM word embedding}\label{alg:tm_train}
\begin{algorithmic}[1]
\Require Vocabulary $\mathcal{V}$; Documents $\mathcal{D} \in \mathcal{G}, \mathcal{D} \subseteq \mathcal{V}$; Accumulation $u$; Clauses $n$; Margin $T$; Specificity $s$; Rounds $r$
\State TMCreate$(n, T, s)$ \Comment{Create TM with $n$ clauses.}
\For{$r$ rounds}
    \For{$\mathit{word}_k \in \mathcal{V}$} \Comment{Create one example per word.}
    \State $q_k \gets \mathrm{Select}(\{0, 1\})$ \Comment{Random target value.}
    \If{$q_k=1$}
    \State $\mathcal{G}_k \gets \{\mathcal{D} | \mathit{word}_k \in \mathcal{D}, \mathcal{D} \in \mathcal{G}\}$ \Comment{Documents with $\mathit{word}_k$.}
    \Else
    \State $\mathcal{G}_k \gets \{\mathcal{D} | \mathit{word}_k \notin \mathcal{D}, \mathcal{D} \in \mathcal{G}\}$ \Comment{Documents without $\mathit{word}_k$.}
    \EndIf
    \State  $\mathcal{S}_k \gets \mathrm{SelectN}(\mathcal{G}_k, u)$ \Comment{Random subset of size $u$.}
    \State $\mathcal{U}_k \gets \bigcup_{\mathcal{D} \in \mathcal{S}_k} \mathcal{D}$ \Comment{Union of selected documents.}
    \State $\mathbold{x}_k \gets (x_1, x_2, \ldots, x_m), x_i = \begin{cases}1,&\mathit{word}_i \in \mathcal{U}^k\\0, &\mathit{word}_i \notin \mathcal{U}^k\end{cases}$
    \State $\mathrm{TMUpdate}(k, \mathbold{x}_k, q_k)$ \Comment{Update TM Autoencoder for output index $k$, input $\mathbold{x}_k$, and  target value $\widehat{x}_k = q_k$.}
    \EndFor
\EndFor
\State $\mathcal{C}, \mathbold{W} \gets \mathrm{TMGetState}()$ \Comment{Clauses $C_j \in \mathcal{C}$ with weights $\mathbold{W}$.}
\State $\mathbold{E} \gets \mathrm{clip}(\mathbold{W}, 0, T)$ \Comment{Elementwise clip of negative values produces weighted logical word embeddings.}
\State $\mathbold{B} \gets \left(\mathbold{W} > 0\right)$ \Comment{Elementwise comparison with zero produces purely logical word embeddings.}
\State \Return $\mathcal{C}, \mathbold{E}, \mathbold{B}$
\end{algorithmic}
\end{algorithm}

\section{Logical Embedding Procedure}
We now use the TM Autoencoder to build logical embeddings. Let $\mathcal{V}= \{\mathit{word}_1, \mathit{word}_2, \ldots, \mathit{word}_m\}$ be the target vocabulary consisting of $m$ unique words.

\paragraph{Pre-processing.} The first step is to pre-process the document corpus. To this end, each document is represented by a subset of words $\mathcal{D} \subseteq \mathcal{V}$. For example, the document ``The actor was brilliant'' becomes the set $\mathcal{D} = \{$``actor'', ``brilliant'', ``the'', ``was''$\}$. The set $\mathcal{G}$, in turn, contains all the documents, $\mathcal{D} \in \mathcal{G}$.  Finally, in propositional vector form, the word set $\mathcal{D}$ becomes:
\begin{equation}
    \mathbold{x} = (x_1, x_2, \ldots, x_t), x_i = \begin{cases}1,&\mathit{word}_i \in D,\\0, &\mathit{word}_i \notin D.\end{cases}
\end{equation}

\paragraph{Embedding.} Algorithm \ref{alg:tm_train} specifies the procedure for embedding the $m$ vocabulary words from $\mathcal{V}$ by using $n$ clauses, $C_j, 1 \le j \le n$, forming a clause set $\mathcal{C}$. Each round of training produces a training example $[k, (x_1, x_2, \ldots, x_m), q_k]$  per $\mathit{word}_k$ in $\mathcal{V}$. First, a target value $q_k$ for the word is set randomly to either $0$ or $1$. This random selection balances the dataset. If $q_k$ becomes $1$, we randomly select $u$ documents that contain $\mathit{word}_k$ and assign them to the set $\mathcal{S}_k$ (positive examples). Otherwise, we randomly select $u$ documents that does \emph{not} contain the word (negative examples). Next, the randomly selected documents are merged by ORing them together, yielding the unified document $\mathcal{U}_k$. The purpose of ORing multiple documents is to increase the frequency of rare context words. Then, picking up characteristic ones becomes easier. After that, the propositional vector form $(x_1, x_2, \ldots, x_m)$ of $\mathcal{U}_k$ is obtained. Finally, the TM Autoencoder is updated with $[k, (x_1, x_2, \ldots, x_m), q_k]$ following the training procedure in Section~\ref{sec:tm_autoencoder}.

\paragraph{Vector Space Representation.} The weighted logical embedding of $\mathit{word}_k \in \mathcal{V}$ can now be obtained from row $k$ of matrix $\mathbold{E}$ (returned from Algorithm \ref{alg:tm_train}), while the the purely logical embedding is found in row $k$ of matrix $\mathbold{B}$. Let $\mathbold{e}_k$ denote the $k$'th row of $\mathbold{E}$, and let $\mathbold{e}_l$ denote the $l$'th row. We can then compare the similarity of two words $\mathit{word}_k$ and $\mathit{word}_l$ using cosine similarity (CS) between their $\mathbold{E}$-embedding:
\begin{equation}
    CS(\mathit{word}_k, \mathit{word}_l) = \frac{\mathbold{e}_k \cdot \mathbold{e}_l}{||\mathbold{e}_k||~ ||\mathbold{e}_l||}.
\end{equation}


\begin{table*}[ht]
\centering
    \begin{adjustbox}{max width=\textwidth}
    \begin{tabular}{l|lll|lll|lll|lll}
    
    \hline
    \multirow{2}{*}{Dataset} &  \multicolumn{3}{c}{W2V} &  \multicolumn{3}{c}{FastText}  &\multicolumn{3}{c}{TM} &  \multicolumn{3}{c}{GloVe}\\ \cline{2-13}
    & Spearman & Kendall& Cosine & Spearman & Kendall & Cosine&Spearman& Kendall & Cosine&Spearman& Kendall & Cosine\\
    \hline
    WS-353     & 0.53 & 0.37 & 0.87  &  0.46 & 0.32 & 0.79  &   0.45 & 0.31 & 0.90 & 0.41 & 0.28 & 0.90  \\
    SIM999 &   0.26 & 0.18 & 0.79  & 0.23 & 0.16 & 0.79  &  0.14 & 0.10 & 0.76  &  0.25 & 0.17 & 0.80  \\
    MEN &   0.71 & 0.50 & 0.91  & 0.71 & 0.51 & 0.94  &  0.64 & 0.45 & 0.94  & 0.73 & 0.53 & 0.95 \\
    MTURK287 &   0.66 & 0.47 & 0.77 & 0.63 & 0.44 & 0.93  &  0.63 & 0.44 & 0.92   & 0.66 & 0.47 & 0.86  \\
    MTURK717 & 0.57 & 0.39 & 0.86  & 0.52 & 0.36 & 0.93 &  0.48 & 0.32 & 0.91 & 0.58 & 0.40 & 0.94 \\
    RG65 & 0.72 & 0.58 & 0.89 & 0.67 & 0.49 & 0.88 & 0.75 & 0.63 & 0.92 & 0.78 & 0.62 & 0.93\\
    \hline
    Average & 0.58 & 0.42 & 0.85 & 0.54 & 0.38 & 0.88 & 0.52 & 0.38 & 0.89 & 0.57 & 0.42 & 0.90\\
    \hline
    \end{tabular}
    \end{adjustbox}
\caption{Performance comparison of TM embedding with baseline algorithms on the similarity task.}
\label{results_similarity}
\end{table*}

\begin{table}[h!!t]
\centering
    \begin{tabular}{l|llll}
    \hline
    Dataset & W2V & FastText & TM & GloVe\\
    \hline
    AP & 0.50 & 0.35 & 0.41 & 0.41\\
    BLESS & 0.64 & 0.66 & 0.62 & 0.66\\
    ESSLI & 0.63 & 0.60 & 0.57 & 0.56\\
    \hline
    Average & 0.59 & 0.54 & 0.53 & 0.54\\
    \hline
    \end{tabular}
\caption{Performance comparison of TM embedding with baseline embeddings on the categorization task.}
\label{results_categorization}
\end{table}

\section{Empirical Evaluation}
\label{evaluation}

We here evaluate our logical embedding scheme, comparing it with neural network approaches.

\subsection{Datasets and Setup}\label{sec:datasets_setup}

 We first evaluate our logical embedding intrinsically, followed by an extrinsic evaluation using classification tasks. 

\paragraph{Intrinsic Evaluation.} We use word similarity and categorization benchmarks for intrinsic evaluation. That is, we examine to what degree our approach retains semantic word relations. To this end, we  measure how semantic relations manifest in vector space using six datasets: SimLex-999, WordSim-353, MEN, MTurk-287, Mturk-771, and RG-65. Each dataset consists of human-scored word pairs, which are compared with the corresponding vector space similarities. The categorization tasks evaluate how well we can group words into distinct word categories, only based on their embedding. We here use three datasets: AP, BLESS, and ESSLLI.\footnote{To obtain the categorization accuracy, we use KMeans clustering from sklearn on the word embeddings and examine the cluster quality by calculating the purity score from (\href{https://github.com/kudkudak/word-embeddings-benchmarks/tree/c78272b8c1374e5e518915a240ab2b348b59f44e}{https://github.com/purity}).} As baselines, we chose Word2Vec, GloVe, and FastText because of their wide use.

\begin{table}[h!!t]
\centering
    \begin{tabular}{l|llllll}
    \hline
    \multirow{2}{*}{Dataset} & \multicolumn{2}{c}{GloVe} & \multicolumn{2}{c}{TM} &\multicolumn{2}{c}{TM$_{\mathrm{hybrid}}$}\\ \cline{2-7}
    & Acc. & F1 & Acc. & F1 & Acc. & F1\\ \hline
    R8 & 96.31 & 0.88 & 96.10 & 0.88 & 97.80 & 0.94\\
    TREC &  95.20 & 0.95 & 96.40 & 0.96 & 96.80 & 0.96\\
    R52 &   90.34 & 0.58 & 91.23 & 0.62 & 94.23 & 0.68\\
    SUBJ & 86.20 & 0.86 & 85.80 & 0.85 & 86.70 & 0.87\\
    SST-2 & 76.38 & 0.75 & 75.61 & 0.74 & 79.30 & 0.78\\
    SST-5 & 47.47 & 0.46 & 47.80 & 0.43 &  49.75 & 0.44\\
    \hline
    \end{tabular}
\caption{Performance comparison of our embedding with standard GloVe embedding on the classification task.}
\label{results_extrinsic}
\end{table}
    
\paragraph{Extrinsic Evaluation.} In our extrinsic evaluation, we investigate how well our logical embedding supports downstream NLP classification tasks. Using the word embeddings as feature vectors, the performance of supervised classification models gives insight into the embedding quality. We employ six standard text classification datasets from SentEval~\cite{conneau2018senteval}: R8, R52, TREC, SUBJ, SST-2, and SST-5. For supervised learning, we use the standard attention-based BiLSTM model with the Adam optimizer and cross-entropy loss function. In this manner, we directly contrast GloVe embedding against the logical TM approach.

\paragraph{Embedding Datasets.} For extrinsic evaluation with BiLSTM, we use standard $300$-dimensional GloVe embeddings, pre-trained on the \textit{Wikipedia 2014 + Gigaword 5} datasets (6B tokens).\footnote{The pre-trained GloVe embeddings can be found here: https://nlp.stanford.edu/projects/glove/} The purpose is to compare the TM embedding performance against widely used and successful GloVe embeddings on downstream tasks. To directly compare the intrinsic properties of Word2Vec, GloVe, FastText, and TM embedding, we also train them from scratch using the One Billion Word dataset~\cite{Chelba2014OneBW}. For training the TM, we use Algorithm \ref{alg:tm_train} with $r=2000$ training rounds, producing $2000$ examples per word by accumulating  $u=25$ contexts per example. We use the following hyperparameters: a pool of $n=600$ clauses, margin  $T=1200$, and specificity $s=5.0$.\footnote{The TM Autoencoder and logical word embedding implementation can be found here: \href{https://github.com/cair/tmu}{https://github.com/cair/tmu}.} Word2Vec Skip-Gram is trained with $10$ passes over the data, using separated embeddings for the input and output contexts. The  window size is $5$ and we use five negative samples per example. Similarly, GloVe is trained for $30$ epochs with a window size of $10$ and a learning rate of $0.05$.\footnote{Word2Vec and FastText have been trained using the standard gensim library (\href{https://github.com/RaRe-Technologies/gensim/tree/develop/gensi}{https://github.com/RaRe-Technologies/gensim/tree/develop/gensi}). GloVe has been trained using \href{https://github.com/maciejkula/glove-python}{https://github.com/maciejkula/glove-python}.}

\begin{figure*}[h!!t]
\centerline{\includegraphics[width=0.7\textwidth]{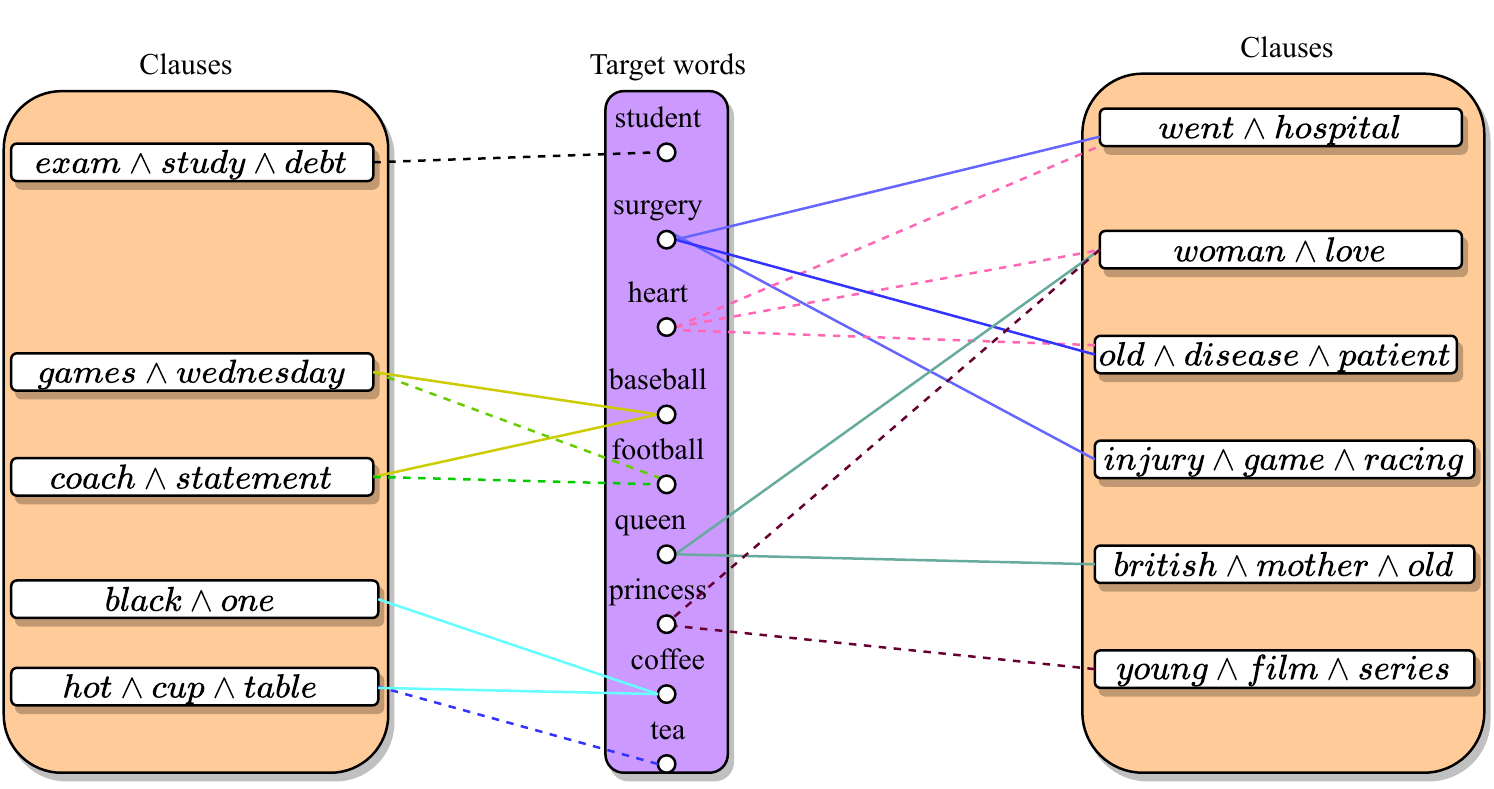}}
\caption{Interpretability of clauses capturing distinct meanings of target words in the TM embedding.}
\label{interpretability_clauses}
\end{figure*}

\subsection{Results and Discussion}
As presented in Section~\ref{sec:datasets_setup}, we employ two kinds of evaluation: intrinsic and extrinsic. Table~\ref{results_similarity} contains the intrinsic evaluation results from the six word similarity tasks. We here compute the Spearman correlation, the Kendall coefficient, and the cosine similarity between the human-set similarity scores and the predicted similarity scores per dataset. Considering Spearman and Kendall score, Word2Vec and GloVe are marginally better than the comparable FastText and TM embedding. However, as reported in~\cite{rastogi2015multiview}, small differences in correlation-based measures are not necessarily significant for smaller datasets.  To more robustly assess performance, we therefore also use cosine similarity to compare predicted word similarities with the human-set similarities. In terms of cosine score, our model outperforms Word2Vec and FastText on the majority of the datasets, while performing competitively with GloVe. This means that the angles between the human-set similarities and the GloVe/TM-predicted similarities are quite similar. Finally, Table~\ref{results_categorization} shows the outcome for the word categorization tasks. As seen, the performance of the selected embedding techniques are comparable, with Word2Vec being slightly ahead.\par

Previous research indicates that intrinsic word similarity performance is minimally or even negatively correlated with downstream NLP performance~\cite{wang2021tsdae}. Therefore, we also include an extrinsic evaluation with six downstream classification tasks. To avoid overfitting and robustly assess downstream properties, we keep our experimental setup from above. Table~\ref{results_extrinsic} reports the outcome of the evaluation, where the embeddings have been fed to an attention-based BiLSTM model. The first configuration (GloVe) uses the pre-trained GloVe embeddings from the \textit{Wikipedia 2014 + Gigaword 5} datasets. The second configuration consists of our purely logical TM embedding from One Billion Word (embedding $\mathbold{B}$ from Algorithm \ref{alg:tm_train}). Being five times smaller, the One Billion Word dataset only provides about $80$ percent of the vocabulary required for the classification tasks. We embed the remaining $20$ percent of the words randomly. Hence, the TM approach can potentially have a disadvantage in the evaluation. In the third configuration (TM$_\mathrm{hybrid}$), we replace the $20$ percent random embeddings with the corresponding GloVe embeddings (approximately 80\% TM + 20\% GloVe). We note that the downstream accuracy of BiLSTM is similar for both TM and GloVe. Specifically, the TM  embedding exceeds GloVe by a small margin on TREC, R52, and SST-5. The hybrid embedding, on the other hand, clearly outperforms the other two. In particular, for R52, SST-2, and SST-5, the hybrid embedding is able to surpass GloVe by a substantial margin of roughly $2-4\%$. Given that the datasets are not completely balanced, we also compute F1 macro scores. We again observe that the TM embedding either outperforms or is competive with GloVe. For R8 and R52, the hybrid embedding surpasses GloVe by a large margin, respectively by around $6\%$ and $10\%$. Based on these results, we  conjecture that logical TM embedding can successfully replace neural network embedding. Even with $20\%$ of the vocabulary missing, trained on five times smaller data, the logical embedding perform competitively with GloVe. Interestingly, the hybrid approach performed even better. One possible explanation of this higher performance can be the extra information added by the larger vocabulary. Additionally, there may be synergy between the neural and logical representations that manifest in the hybrid approach.

\subsection{Interpretability and Visualization}
In this section, we investigate the nature of the TM embeddings in more detail, focusing on interpretability. Our embedding consists of the positive clause weights $\mathbold{E}$, or, alternatively, the propositional version $\mathbold{B}$, explained by the set of clauses $\mathcal{C}$. As demonstrated in Figure~\ref{interpretability_clauses}, each clause in $\mathcal{C}$ captures a facet of a context. The dotted lines in the figure showcase the connection between the target words and their clauses from matrix $\mathbold{B}$ (and, accordingly,  $\mathbold{E}$). Each target word gets its own color to more easily discern the connections. In the figure, we provide an excerpt of $18$ connections from  $\mathbold{B}$, involving $8$  target words and the $11$ most triggered clauses for these words. Consider for example the target words \emph{surgery} and \emph{heart}. These two target words share two clauses: [$\mathit{went}~\wedge ~\mathit{hospital}$] and [$\mathit{old}~\wedge~\mathit{disease}~\wedge~{patient}$]. The two clauses capture two joint contexts, both related to health. The clauses thus represent commonality between the target words, providing information on one particular meaning of the words.

The two target words are also semantically different. The differences are captured by the clauses they do not share. The target word \emph{heart}, for example, also relates to the meaning [$\mathit{woman}~\wedge~\mathit{love}$], which \emph{surgery} does not. \emph{Surgery}, on the other hand, connects with [$\mathit{injury}~\wedge~\mathit{game}~\wedge~\mathit{racing}$]. In this manner, the unique meanings and relations between words are represented through sharing of logical expressions. Accordingly, it is feasible to capture a wide range of possible contextual representations with concise logical expressions. As such, the logical embedding provides a sparse representation of words and their relations. Indeed, at most $10\%$ of the clauses connect to each word in our experiments.  As shown in the intrinsic evaluations from the previous subsection, these contextual representations are effective for measuring word similarity and categorizing words. Similarly, we observed that the logical embedding is boosting downstream NLP classification tasks.

\begin{figure}
\centerline{\includegraphics[width=0.5\textwidth]{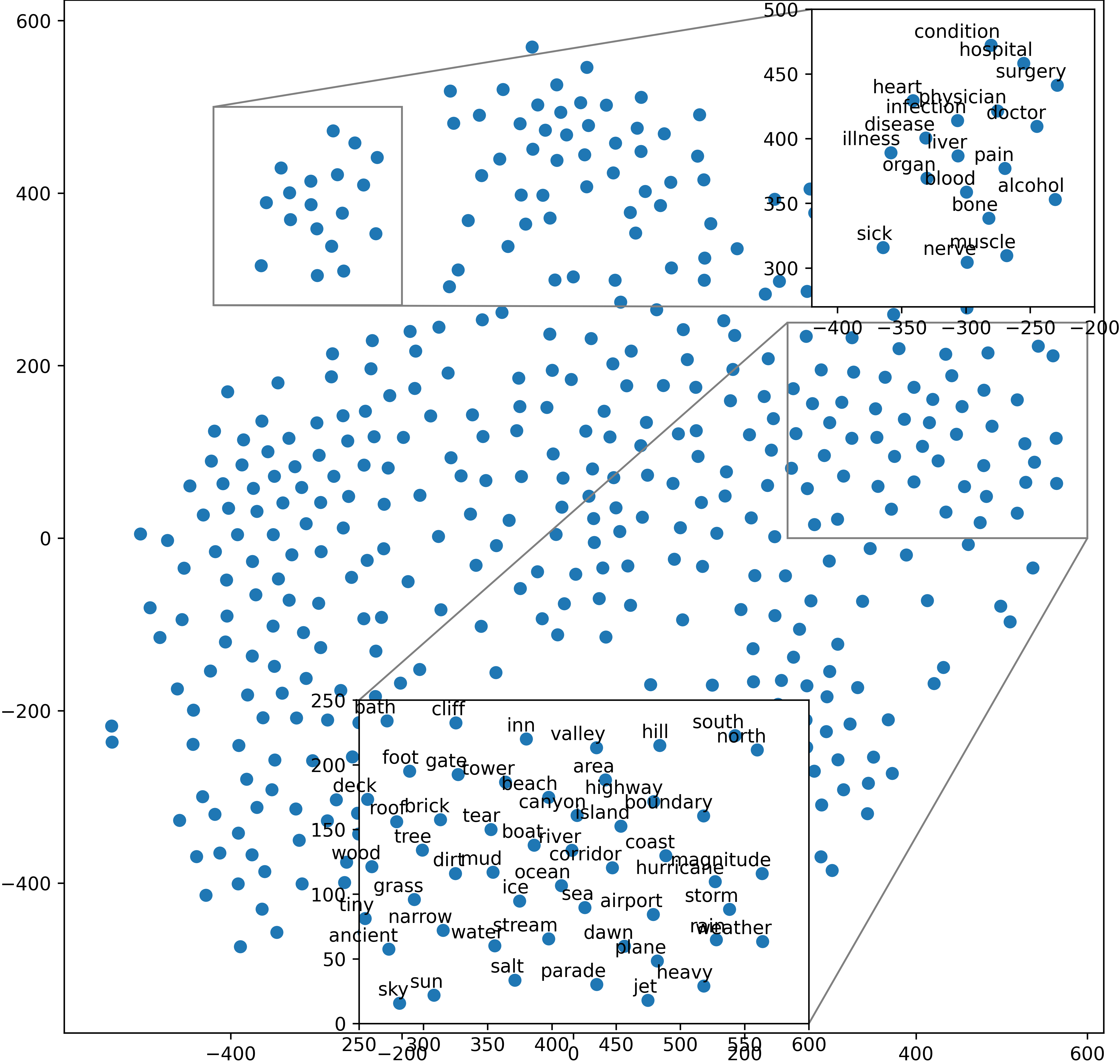}}
\caption{TM embedding visualization plotted using t-SNE.}
\label{visualization_target_word}
\end{figure}

\par
To cast further light on the TM embedding approach, we visualize the embedding of $400$ words from the SimLex-999 dataset in Figure~\ref{visualization_target_word}, plotted using t-SNE. The figure indicates that we are able to cluster contextually similar words in vector space. To scrutinize the clusters, we zoom in on two of them. Consider the upper right cluster first. Notice how the words in the cluster relate to \emph{hospital}, such as \emph{heart} and \emph{diseases}. As seen, the word embeddings are closely located in vector space. Similarly, we can observe that terminology connected to weather and geography are grouped together in the bottom cluster. From these two examples, it seems clear that the TM embedding incorporates semantic relationships among words.

\section{Conclusion and Future Work}
In this work, we first discussed the challenge and necessity of finding computationally simpler and more interpretable word embedding approaches. We then motivated an efficient self-supervised approach, namely, a TM-based autoencoder, for producing sparse and interpretable logical word embeddings.  
We evaluated our approach on a wide range of intrinsic and extrinsic tasks, demonstrating that it is competitive with dense neural network-based embedding schemes such as Word2Vec, GloVe, and FastText. Further, we investigated the interpretability our embedding through visualization and a case study. Our conclusion from the study is that the logical embedding is able to represent words with logical expressions. This structure makes the representation sparse, enabling a clear-cut decomposition of each word into sets of semantic concepts. \par
Future work includes scaling up our implementation using GPUs to support building of large scale vocabularies from more massive datasets. Also, we intend to investigate how sentence-level and document-level embedding can be created using clauses, for instance applicable for downstream sentence similarity tasks.


\bibliographystyle{named}
\bibliography{ijcai22}

\begin{thebibliography}{}

\bibitem[\protect\citeauthoryear{Angelov}{2020}]{angelov2020top2vec}
Dimo Angelov.
\newblock {Top2vec: Distributed representations of topics}.
\newblock {\em arXiv preprint arXiv:2008.09470}, 2020.

\bibitem[\protect\citeauthoryear{Arora \bgroup \em et al.\egroup
  }{2017}]{arora2017simple}
Sanjeev Arora, Yingyu Liang, and Tengyu Ma.
\newblock {A simple but tough-to-beat baseline for sentence embeddings}.
\newblock In {\em International conference on learning representations}, 2017.

\bibitem[\protect\citeauthoryear{Arora \bgroup \em et al.\egroup
  }{2018}]{arora2018linear}
Sanjeev Arora, Yuanzhi Li, Yingyu Liang, Tengyu Ma, and Andrej Risteski.
\newblock {Linear algebraic structure of word senses, with applications to
  polysemy}.
\newblock {\em Transactions of the Association for Computational Linguistics},
  6:483--495, 2018.

\bibitem[\protect\citeauthoryear{Bengio \bgroup \em et al.\egroup
  }{2000}]{bengio2000neural}
Yoshua Bengio, R{\'e}jean Ducharme, and Pascal Vincent.
\newblock {A neural probabilistic language model}.
\newblock {\em Advances in neural information processing systems}, 13, 2000.

\bibitem[\protect\citeauthoryear{Bhattarai \bgroup \em et al.\egroup
  }{2022a}]{bhattarai2022tsetlin}
Bimal Bhattarai, Ole-Christoffer Granmo, and Lei Jiao.
\newblock {A Tsetlin Machine Framework for Universal Outlier and Novelty
  Detection}.
\newblock In {\em International Conference on Agents and Artificial
  Intelligence}, pages 250--268. Springer, 2022.

\bibitem[\protect\citeauthoryear{Bhattarai \bgroup \em et al.\egroup
  }{2022b}]{8}
Bimal Bhattarai, Ole-Christoffer Granmo, and Lei Jiao.
\newblock {Explainable Tsetlin Machine framework for fake news detection with
  credibility score assessment}.
\newblock In {\em LREC}, 2022.

\bibitem[\protect\citeauthoryear{{Blakely} and
  {Granmo}}{2021}]{blakely2021closed}
Christian~D. {Blakely} and Ole-Christoffer {Granmo}.
\newblock {Closed-Form Expressions for Global and Local Interpretation of
  Tsetlin Machines}.
\newblock In {\em 34th International Conference on Industrial, Engineering and
  Other Applications of Applied Intelligent Systems (IEA/AIE 2021)}. Springer,
  2021.

\bibitem[\protect\citeauthoryear{Bojanowski \bgroup \em et al.\egroup
  }{2017}]{bojanowski2017enriching}
Piotr Bojanowski, Edouard Grave, Armand Joulin, and Tomas Mikolov.
\newblock {Enriching word vectors with subword information}.
\newblock {\em Transactions of the association for computational linguistics},
  5:135--146, 2017.

\bibitem[\protect\citeauthoryear{Borgeaud \bgroup \em et al.\egroup
  }{2022}]{borgeaud2022improving}
Sebastian Borgeaud, Arthur Mensch, Jordan Hoffmann, Trevor Cai, Eliza
  Rutherford, Katie Millican, George~Bm Van Den~Driessche, Jean-Baptiste
  Lespiau, Bogdan Damoc, Aidan Clark, et~al.
\newblock {Improving language models by retrieving from trillions of tokens}.
\newblock In {\em International conference on machine learning}, pages
  2206--2240. PMLR, 2022.

\bibitem[\protect\citeauthoryear{Cer \bgroup \em et al.\egroup
  }{2018}]{cer2018universal}
Daniel Cer, Yinfei Yang, Sheng-yi Kong, Nan Hua, Nicole Limtiaco, Rhomni~St
  John, Noah Constant, Mario Guajardo-Cespedes, Steve Yuan, Chris Tar, et~al.
\newblock {Universal sentence encoder for English}.
\newblock In {\em Proceedings of the 2018 conference on empirical methods in
  natural language processing: system demonstrations}, pages 169--174, 2018.

\bibitem[\protect\citeauthoryear{Charniak and others}{2013}]{charniak2013naive}
Eugene Charniak et~al.
\newblock {Naive Bayes word sense induction}.
\newblock In {\em Proceedings of the 2013 Conference on Empirical Methods in
  Natural Language Processing}, pages 1433--1437, 2013.

\bibitem[\protect\citeauthoryear{Chelba \bgroup \em et al.\egroup
  }{2014}]{Chelba2014OneBW}
Ciprian Chelba, Tomas Mikolov, Mike Schuster, Qi~Ge, T.~Brants, Phillip~Todd
  Koehn, and Tony Robinson.
\newblock One billion word benchmark for measuring progress in statistical
  language modeling.
\newblock In {\em Interspeech}, 2014.

\bibitem[\protect\citeauthoryear{Conneau and Kiela}{2018}]{conneau2018senteval}
Alexis Conneau and Douwe Kiela.
\newblock {SentEval: An Evaluation Toolkit for Universal Sentence
  Representations}.
\newblock In {\em Proceedings of the Eleventh International Conference on
  Language Resources and Evaluation (LREC 2018)}, 2018.

\bibitem[\protect\citeauthoryear{Devlin \bgroup \em et al.\egroup
  }{2019}]{Devlin2019BERTPO}
Jacob Devlin, Ming-Wei Chang, Kenton Lee, and Kristina Toutanova.
\newblock {BERT: Pre-training of Deep Bidirectional Transformers for Language
  Understanding}.
\newblock In {\em NAACL}, 2019.

\bibitem[\protect\citeauthoryear{Dhillon \bgroup \em et al.\egroup
  }{2015}]{dhillon2015eigenwords}
Paramveer~S Dhillon, Dean~P Foster, and Lyle~H Ungar.
\newblock {Eigenwords: spectral word embeddings}.
\newblock {\em J. Mach. Learn. Res.}, 16:3035--3078, 2015.

\bibitem[\protect\citeauthoryear{Faruqui \bgroup \em et al.\egroup
  }{2015}]{faruqui2015sparse}
Manaal Faruqui, Yulia Tsvetkov, Dani Yogatama, Chris Dyer, and Noah~A Smith.
\newblock {Sparse Overcomplete Word Vector Representations}.
\newblock In {\em Proceedings of the 53rd Annual Meeting of the Association for
  Computational Linguistics and the 7th International Joint Conference on
  Natural Language Processing (Volume 1: Long Papers)}, pages 1491--1500, 2015.

\bibitem[\protect\citeauthoryear{Glimsdal and
  Granmo}{2021}]{glimsdal2021coalesced}
Sondre Glimsdal and Ole-Christoffer Granmo.
\newblock {Coalesced Multi-Output Tsetlin Machines with Clause Sharing}.
\newblock {\em Submitted to ICML'23. Available as an arXiv preprint,
  arXiv:2108.07594}, 2021.

\bibitem[\protect\citeauthoryear{{Granmo}}{2018}]{granmo2018tsetlin}
Ole-Christoffer {Granmo}.
\newblock {The Tsetlin Machine - A Game Theoretic Bandit Driven Approach to
  Optimal Pattern Recognition with Propositional Logic}.
\newblock {\em arXiv preprint arXiv:1804.01508}, 2018.

\bibitem[\protect\citeauthoryear{Harris}{1954}]{harris1954distributional}
Zellig~S Harris.
\newblock {Distributional structure}.
\newblock {\em Word}, 10(2-3):146--162, 1954.

\bibitem[\protect\citeauthoryear{Huang \bgroup \em et al.\egroup
  }{2020}]{huang2020embedding}
Jui-Ting Huang, Ashish Sharma, Shuying Sun, Li~Xia, David Zhang, Philip Pronin,
  Janani Padmanabhan, Giuseppe Ottaviano, and Linjun Yang.
\newblock {Embedding-based retrieval in Facebook search}.
\newblock In {\em Proceedings of the 26th ACM SIGKDD International Conference
  on Knowledge Discovery \& Data Mining}, pages 2553--2561, 2020.

\bibitem[\protect\citeauthoryear{Kenton and Toutanova}{2019}]{kenton2019bert}
Jacob Devlin Ming-Wei~Chang Kenton and Lee~Kristina Toutanova.
\newblock Bert: Pre-training of deep bidirectional transformers for language
  understanding.
\newblock In {\em Proceedings of NAACL-HLT}, pages 4171--4186, 2019.

\bibitem[\protect\citeauthoryear{Lample \bgroup \em et al.\egroup
  }{2016}]{lample2016neural}
Guillaume Lample, Miguel Ballesteros, Sandeep Subramanian, Kazuya Kawakami, and
  Chris Dyer.
\newblock {Neural Architectures for Named Entity Recognition}.
\newblock In {\em Proceedings of NAACL-HLT}, pages 260--270, 2016.

\bibitem[\protect\citeauthoryear{Lau \bgroup \em et al.\egroup
  }{2014}]{lau2014learning}
Jey~Han Lau, Paul Cook, Diana McCarthy, Spandana Gella, and Timothy Baldwin.
\newblock Learning word sense distributions, detecting unattested senses and
  identifying novel senses using topic models.
\newblock In {\em Proceedings of the 52nd Annual Meeting of the Association for
  Computational Linguistics (Volume 1: Long Papers)}, pages 259--270, 2014.

\bibitem[\protect\citeauthoryear{Levy \bgroup \em et al.\egroup
  }{2015}]{levy2015improving}
Omer Levy, Yoav Goldberg, and Ido Dagan.
\newblock {Improving distributional similarity with lessons learned from word
  embeddings}.
\newblock {\em Transactions of the association for computational linguistics},
  3:211--225, 2015.

\bibitem[\protect\citeauthoryear{Li \bgroup \em et al.\egroup
  }{2020}]{li2020sentence}
Bohan Li, Hao Zhou, Junxian He, Mingxuan Wang, Yiming Yang, and Lei Li.
\newblock {On the Sentence Embeddings from Pre-trained Language Models}.
\newblock In {\em Proceedings of the 2020 Conference on Empirical Methods in
  Natural Language Processing (EMNLP)}, pages 9119--9130, 2020.

\bibitem[\protect\citeauthoryear{Logeswaran and
  Lee}{2018}]{logeswaran2018efficient}
Lajanugen Logeswaran and Honglak Lee.
\newblock {An efficient framework for learning sentence representations}.
\newblock In {\em International Conference on Learning Representations}, 2018.

\bibitem[\protect\citeauthoryear{Mikolov \bgroup \em et al.\egroup
  }{2013}]{Mikolov2013EfficientEO}
Tomas Mikolov, Kai Chen, Gregory~S. Corrado, and Jeffrey Dean.
\newblock {Efficient Estimation of Word Representations in Vector Space}.
\newblock In {\em International Conference on Learning Representations}, 2013.

\bibitem[\protect\citeauthoryear{Mu \bgroup \em et al.\egroup
  }{2017}]{mu2017geometry}
Jiaqi Mu, Suma Bhat, and Pramod Viswanath.
\newblock Geometry of polysemy.
\newblock In {\em 5th International Conference on Learning Representations,
  ICLR 2017}, 2017.

\bibitem[\protect\citeauthoryear{Pennington \bgroup \em et al.\egroup
  }{2014}]{pennington2014glove}
Jeffrey Pennington, Richard Socher, and Christopher~D Manning.
\newblock {Glove: Global vectors for word representation}.
\newblock In {\em Proceedings of the 2014 conference on empirical methods in
  natural language processing (EMNLP)}, pages 1532--1543, 2014.

\bibitem[\protect\citeauthoryear{Peters \bgroup \em et al.\egroup
  }{2018}]{peters1802deep}
Matthew~E Peters, Mark Neumann, Mohit Iyyer, Matt Gardner, Christopher Clark,
  Kenton Lee, and Luke Zettlemoyer.
\newblock {Deep contextualized word representations}.
\newblock {\em arXiv preprint arXiv:1802.05365}, 2018.

\bibitem[\protect\citeauthoryear{Rastogi \bgroup \em et al.\egroup
  }{2015}]{rastogi2015multiview}
Pushpendre Rastogi, Benjamin Van~Durme, and Raman Arora.
\newblock {Multiview LSA: Representation learning via generalized CCA}.
\newblock In {\em Proceedings of the 2015 conference of the North American
  chapter of the Association for Computational Linguistics: human language
  technologies}, pages 556--566, 2015.

\bibitem[\protect\citeauthoryear{Reimers and
  Gurevych}{2019}]{reimers2019sentence}
Nils Reimers and Iryna Gurevych.
\newblock {Sentence-BERT: Sentence Embeddings using Siamese BERT-Networks}.
\newblock In {\em Proceedings of the 2019 Conference on Empirical Methods in
  Natural Language Processing and the 9th International Joint Conference on
  Natural Language Processing (EMNLP-IJCNLP)}, pages 3982--3992, 2019.

\bibitem[\protect\citeauthoryear{Schwartz \bgroup \em et al.\egroup
  }{2020}]{Schwartz2020GreenA}
Roy Schwartz, Jesse Dodge, Noah Smith, and Oren Etzioni.
\newblock {Green AI}.
\newblock {\em Communications of the ACM}, 63:54 -- 63, 2020.

\bibitem[\protect\citeauthoryear{Subramanian \bgroup \em et al.\egroup
  }{2018}]{subramanian2018spine}
Anant Subramanian, Danish Pruthi, Harsh Jhamtani, Taylor Berg-Kirkpatrick, and
  Eduard Hovy.
\newblock {Spine: Sparse interpretable neural embeddings}.
\newblock In {\em Proceedings of the AAAI Conference on Artificial
  Intelligence}, volume~32, 2018.

\bibitem[\protect\citeauthoryear{Vaswani \bgroup \em et al.\egroup
  }{2017}]{vaswani2017attention}
Ashish Vaswani, Noam Shazeer, Niki Parmar, Jakob Uszkoreit, Llion Jones,
  Aidan~N Gomez, {\L}ukasz Kaiser, and Illia Polosukhin.
\newblock {Attention is all you need}.
\newblock {\em Advances in neural information processing systems}, 30, 2017.

\bibitem[\protect\citeauthoryear{Wang \bgroup \em et al.\egroup
  }{2021}]{wang2021tsdae}
Kexin Wang, Nils Reimers, and Iryna Gurevych.
\newblock {Tsdae: Using transformer-based sequential denoising auto-encoder for
  unsupervised sentence embedding learning}.
\newblock {\em arXiv preprint arXiv:2104.06979}, 2021.

\bibitem[\protect\citeauthoryear{Yadav \bgroup \em et al.\egroup }{2021}]{6}
Rohan Yadav, Lei Jiao, Ole-Christoffer Granmo, and Morten Goodwin.
\newblock {Human-Level Interpretable Learning for Aspect-Based Sentiment
  Analysis}.
\newblock In {\em Proceedings of AAAI}, 2021.

\bibitem[\protect\citeauthoryear{Zhang \bgroup \em et al.\egroup
  }{2020}]{zhang2020unsupervised}
Yan Zhang, Ruidan He, Zuozhu Liu, Kwan~Hui Lim, and Lidong Bing.
\newblock {An Unsupervised Sentence Embedding Method by Mutual Information
  Maximization}.
\newblock In {\em Proceedings of the 2020 Conference on Empirical Methods in
  Natural Language Processing (EMNLP)}, pages 1601--1610, 2020.

\end{thebibliography}

\end{document}